\documentclass[11pt,a4paper]{article}
\usepackage[hyperref]{acl2020}
\usepackage{times}
\usepackage{latexsym}

\usepackage{graphicx}
\usepackage{amsmath}
\usepackage{amsfonts}
\usepackage{multirow}

\usepackage{microtype}

\aclfinalcopy 
\def\aclpaperid{1515} 

\title{Leveraging Graph to Improve Abstractive Multi-Document Summarization}

\author{Wei Li\textsuperscript{1}, Xinyan Xiao\textsuperscript{1}\thanks{Corresponding author.}, Jiachen Liu\textsuperscript{1}, Hua Wu\textsuperscript{1}, Haifeng Wang\textsuperscript{1}, Junping Du\textsuperscript{2} \\
  \textsuperscript{1}Baidu Inc., Beijing, China \\
  \textsuperscript{2}Beijing University of Posts and Telecommunications \\
  \texttt{\{liwei85,xiaoxinyan,liujiachen,wu\_hua,wanghaifeng\}@baidu.com}\\
  \texttt{junpingdu@126.com}
  }

\date{}

\begin{document}
\maketitle
\begin{abstract}
Graphs that capture relations between textual units have great benefits for detecting salient information from multiple documents and generating overall coherent summaries.
In this paper, we develop a neural abstractive multi-document summarization (MDS) model which can leverage well-known graph representations of documents such as similarity graph and discourse graph, to more effectively process multiple input documents and produce abstractive summaries.
Our model utilizes graphs to encode documents in order to capture cross-document relations, which is crucial to summarizing long documents.
Our model can also take advantage of graphs to guide the summary generation process, which is beneficial for generating coherent and concise summaries.
Furthermore, pre-trained language models can be easily combined with our model, which further improve the summarization performance significantly.
Empirical results on the WikiSum and MultiNews dataset show that the proposed architecture brings substantial improvements over several strong baselines.
\end{abstract}

\section{Introduction}

Multi-document summarization (MDS) brings great challenges to the widely used sequence-to-sequence (Seq2Seq) neural architecture as it requires effective representation of multiple input documents and content organization of long summaries.
For MDS, different documents may contain the same content, include additional information, and present complementary or contradictory information \citep{radev2000common}.
So different from single document summarization (SDS), cross-document links are very important in extracting salient information, detecting redundancy and generating overall coherent summaries for MDS.
Graphs that capture relations between textual units have great benefits to MDS, which can help generate more informative, concise and coherent summaries from multiple documents.
Moreover, graphs can be easily constructed by representing text spans (e.g. sentences, paragraphs etc.) as graph nodes and the semantic links between them as edges.
Graph representations of documents such as similarity graph based on lexical similarities \citep{erkan2004lexrank} and discourse graph based on discourse relations \citep{christensen2013towards}, have been widely used in traditional graph-based extractive MDS models.
However, they are not well studied by most abstractive approaches, especially the end-to-end neural approaches.
Few work has studied the effectiveness of explicit graph representations on neural abstractive MDS.

In this paper, we develop a neural abstractive MDS model which can leverage explicit graph representations of documents to more effectively process multiple input documents and distill abstractive summaries.
Our model augments the end-to-end neural architecture with the ability to incorporate well-established graphs into both the document representation and summary generation processes.
Specifically, a graph-informed attention mechanism is developed to incorporate graphs into the document encoding process, which enables our model to capture richer cross-document relations.
Furthermore, graphs are utilized to guide the summary generation process via a hierarchical graph attention mechanism, which takes advantage of the explicit graph structure to help organize the summary content.
Benefiting from the graph modeling, our model can extract salient information from long documents and generate coherent summaries more effectively.
We experiment with three types of graph representations, including similarity graph, topic graph and discourse graph, which all significantly improve the MDS performance.

Additionally, our model is complementary to most pre-trained language models (LMs), like BERT \citep{devlin2018bert}, RoBERTa \citep{liu2019roberta} and XLNet \citep{yang2019xlnet}.
They can be easily combined with our model to process much longer inputs.
The combined model adopts the advantages of both our graph model and pre-trained LMs.
Our experimental results show that our graph model significantly improves the performance of pre-trained LMs on MDS.

The contributions of our paper are as follows:
\begin{itemize}
    \item Our work demonstrates the effectiveness of graph modeling in neural abstractive MDS. We show that explicit graph representations are beneficial for both document representation and summary generation.
	\item We propose an effective method to incorporate explicit graph representations into the neural architecture, and an effective method to combine pre-trained LMs with our graph model to process long inputs more effectively.
	\item Our model brings substantial improvements over several strong baselines on both WikiSum and MultiNews dataset. We also report extensive analysis results, demonstrating that graph modeling enables our model process longer inputs with better performance, and graphs with richer relations are more beneficial for MDS.\footnote{Codes and results are in: \url{https://github.com/PaddlePaddle/Research/tree/master/NLP/ACL2020-GraphSum}}
\end{itemize}

\section{Related Work}

\subsection{Graph-based MDS}
\label{ssect:rel_graph}
Most previous MDS approaches are extractive, which extract salient textual units from documents based on graph-based representations of sentences.
Various ranking methods have been developed to rank textual units based on graphs to select most salient ones for inclusion in the final summary.

\citet{erkan2004lexrank} propose LexRank to compute sentence importance based on a lexical similarity graph of sentences.
\citet{mihalcea2004textrank} propose a graph-based ranking model to extract salient sentences from documents.
\citet{wan2008exploration} further proposes to incorporate document-level information and sentence-to-document relations into the graph-based ranking process.
A series of variants of the PageRank algorithm has been further developed to compute the salience of textual units recursively based on various graph representations of documents \citep{wan2009graph, cai2012mutually}.
More recently, \citet{yasunaga2017graph} propose a neural graph-based model for extractive MDS. 
An approximate discourse graph is constructed based on discourse markers and entity links.
The salience of sentences is estimated using features from graph convolutional networks \citep{kipf2016semi}.
\citet{yin2019graph} also propose a graph-based neural sentence ordering model, which utilizes entity linking graph to capture the global dependencies between sentences.

\subsection{Abstractive MDS}
\label{ssect:rel_abs}

Abstractive MDS approaches have met with limited success.
Traditional approaches mainly include: sentence fusion-based
\citep{banerjee2015multi, filippova2008sentence, barzilay2005sentence, barzilay2003information},
information extraction-based \citep{li2015abstractive, pighin2014modelling, wang2013domain, genest2011framework, li2019abstractive}
and paraphrasing-based \citep{bing2015abstractive, berg2011jointly, cohn2009sentence}.
More recently, some researches parse the source text into AMR representation and then generate summary based on it \citep{liao2018abstract}.

\begin{figure*}[t!]
	\centering
	\includegraphics[width=5in]{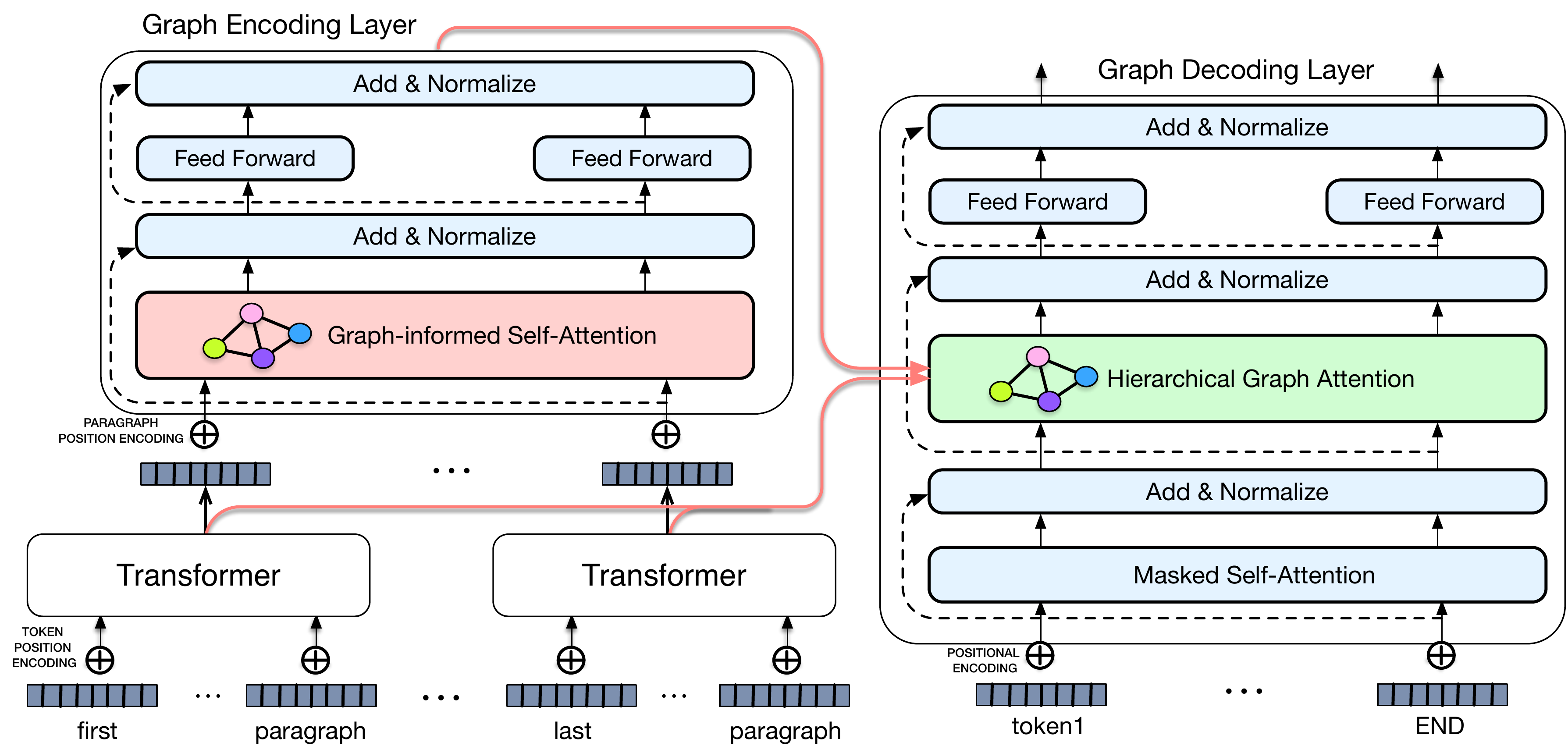}
	\caption{Illustration of our model, which follows the encoder-deocder architecture. The encoder is a stack of transformer layers and graph encoding layers, while the decoder is a stack of graph decoding layers. We incorporate explicit graph representations into both the graph encoding layers and graph decoding layers.}
	\label{fig:graphsumm}
\end{figure*}

Although neural abstractive models have achieved promising results on SDS \citep{see2017get, paulus2018deep, gehrmann2018bottom, celikyilmaz2018deep, li2018improving, li2018improving2, narayan2018don, yang2019interactive, sharma2019entity, perez2019generating}, it's not straightforward to extend them to MDS.
Due to the lack of sufficient training data, earlier approaches try to simply transfer SDS model to MDS task \citep{lebanoff2018adapting, zhang2018towards, baumel2018query} or utilize unsupervised models relying on reconstruction objectives \citep{ma2016unsupervised, chu2019meansum}.
Later, \citet{liu2018generating} propose to construct a large scale MDS dataset (namely WikiSum) based on Wikipedia, and develop a Seq2Seq model by considering the multiple input documents as a concatenated flat sequence.
\citet{fan2019using} further propose to construct a local knowledge graph from documents and then linearize the graph into a sequence to better sale Seq2Seq models to multi-document inputs. 
\citet{fabbri2019multi} also introduce a middle-scale (about 50K) MDS news dataset (namely MultiNews), and propose an end-to-end model by incorporating traditional MMR-based extractive model with a standard Seq2Seq model. 
The above Seq2Seq models haven't study the importance of cross-document relations and graph representations in MDS.

Most recently, \citet{liu2019hierarchical} propose a hierarchical transformer model to utilize the hierarchical structure of documents. 
They propose to learn cross-document relations based on  self-attention mechanism.
They also propose to incorporate explicit graph representations into the model by simply replacing the attention weights with a graph matrix, however, it doesn't achieve obvious improvement according to their experiments.
Our work is partly inspired by this work, but our approach is quite different from theirs. 
In contrast to their approach, we incorporate explicit graph representations into the encoding process via a graph-informed attention mechanism. 
Under the guidance of explicit relations in graphs, our model can learn better and richer cross-document relations, thus achieves significantly better performance.
We also leverage the graph structure to guide the summary decoding process, which is beneficial for long summary generation.
Additionally, we combine the advantages of pretrained LMs into our model.

\subsection{Summarization with Pretrained LMs}
\label{ssect:rel_lm}

Pretrained LMs \citep{peters2018deep, radford2018improving, devlin2018bert, dong2019unified, sun2019ernie} have recently emerged as a key technology for achieving impressive improvements in a wide variety of natural language tasks, including both language understanding and language generation \citep{edunov2019pre, rothe2019leveraging}.
\citet{liu2019text} attempt to incorporate pre-trained BERT encoder into SDS model and achieves significant improvements.
\citet{dong2019unified} further propose a unified LM for both language understanding and language generation tasks, which achieves state-of-the-art results on several generation tasks including SDS.
In this work, we propose an effective method to combine pretrained LMs with our graph model and make them be able to process much longer inputs effectively.

\section{Model Description}
\label{sect:model}
In order to process long source documents more effectively, we follow \citet{liu2019hierarchical} in splitting source documents into multiple paragraphs by line-breaks. 
Then the graph representation of documents is constructed over paragraphs.
For example, a similarity graph can be built based on cosine similarities between tf-idf representations of paragraphs.
Let $\mathbb{G}$ denotes a graph representation matrix of the input documents, where $\mathbb{G}[i][j]$ indicates the relation weights between paragraph $P_i$ and $P_j$.
Formally, the task is to generate the summary $S$ of the document collection given $L$ input paragraphs ${P_1,\ldots,P_L}$ and their graph representation $\mathbb{G}$.

Our model is illustrated in Figure~\ref{fig:graphsumm}, which follows the encoder-decoder architecture \citep{bahdanau2015neural}.
The encoder is composed of several token-level transformer encoding layers and paragraph-level graph encoding layers which can be stacked freely.
The transformer encoding layer follows the Transformer architecture introduced in \citet{vaswani2017attention}, encoding contextual information for tokens within each paragraph.
The graph encoding layer extends the Transformer architecture with a graph attention mechanism to incorporate explicit graph representations into the encoding process.
Similarly, the decoder is composed of a stack of graph decoding layers.
They extend the Transformer with a hierarchical graph attention mechanism to utilize explicit graph structure to guide the summary decoding process.
In the following, we will focus on the graph encoding layer and graph decoding layer of our model.

\subsection{Graph Encoding Layer}
\label{ssect:model_enc}
As shown in Figure \ref{fig:graphsumm}, based on the output of the token-level transformer encoding layers, the graph encoding layer is used to encode all documents globally.
Most existing neural work only utilizes attention mechanism to learn latent graph representations of documents where the graph edges are attention weights \citep{liu2019hierarchical, niculae2018towards, fernandes2018structured}.
However, much work in traditional MDS has shown that explicit graph representations are very beneficial to MDS.
Different types of graphs capture different kinds of semantic relations (e.g. lexical relations or discourse relations), which can help the model focus on different facets of the summarization task.
In this work, we propose to incorporate explicit graph representations into the neural encoding process via a graph-informed attention mechanism.
It takes advantage of the explicit relations in graphs to learn better inter-paragraph relations.
Each paragraph can collect information from other related paragraphs to capture global information from the whole input.

\paragraph{Graph-informed Self-attention} 
The graph-informed self-attention extends the self-attention mechanism to consider the pairwise relations in explicit graph representations.
Let $x_{i}^{l-1}$ denotes the output of the $(l-1)$-th graph encoding layer for paragraph $P_i$, where $x_{i}^{0}$ is just the input paragraph vector.
For each paragraph $P_i$, 
the context representation $u_{i}$ can be computed as a weighted sum of linearly transformed paragraph vectors:
\begin{equation}
\begin{aligned}
    \alpha_{ij} =& softmax(e_{ij} + \Re_{ij}) \\
    e_{ij} =& \frac{(x_{i}^{l-1}W_{Q})(x_{j}^{l-1}W_{K})^T}{\sqrt{d_{head}}}  \\
	u_{i} =& \sum_{j=1}^{L} \alpha_{ij}(x_{j}^{l-1}W_{V})
\end{aligned}
\label{eq3}
\end{equation}
\noindent where $W_{K}$, $W_{Q}$ and $W_{V} \in \mathbb{R}^{d*d}$ are parameter weights. $e_{tj}$ denotes the latent relation weight between paragraph $P_{i}$ and $P_{j}$.
The main difference of our graph-informed self-attention is the additional pairwise relation bias $\Re_{ij}$, which is computed as a Gaussian bias of the weights of graph representation matrix $\mathbb{G}$:
\begin{equation}
\begin{aligned}
    \Re_{ij} = -\frac{({1 - \mathbb{G}[i][j]})^{2}}{2{\sigma}^{2}}
\end{aligned}
\label{eq4}
\end{equation}
where $\sigma$ denotes the standard deviation that represents the influence intensity of the graph structure. We set it empirically by tuning on the development dataset. The gaussian bias $R_{ij} \in (-inf, 0]$ measures the tightness between the paragraphs $P_i$ and $P_j$. Due to the exponential operation in softmax function, the gaussian bias approximates to multiply the latent attention distribution by a weight $\in (0, 1]$.

In our graph-attention mechanism, the term $e_{ij}$ in Equation~\ref{eq3} keeps the ability to model latent dependencies between any two paragraphs, and the term $\Re_{ij}$ incorporates explicit graph representations as prior constraints into the encoding process. This way, our model can learn better and richer inter-paragraph relations to obtain more informative paragraph representations.


Then, a two-layer feed-forward network with ReLU activation function and a high-way layer normalization are applied to obtain the vector of each paragraph $x_{i}^{l}$:
\begin{equation}
\begin{aligned}
    p_i^{l} =& W_{o2}ReLU(W_{o1}(u_{i} + x_{i}^{l-1})) \\
    x_{i}^{l} =& LayerNorm(p_i^{l} + x_{i}^{l-1})
\end{aligned}
\label{eq6}
\end{equation}
where $W_{o1} \in \mathbb{R}^{d_{ff}*d}$ and $W_{o2} \in \mathbb{R}^{d*d_{ff}}$ are learnable parameters, $d_{ff}$ is the hidden size of the feed-forward layer.

\subsection{Graph Decoding Layer}
\label{ssect:model_dec}
Graphs can also contribute to the summary generation process.
The relations between textual units can help to generate more coherent or concise summaries. 
For example, \citet{christensen2013towards} propose to leverage an approximate discourse graph to help generate coherent extractive summaries. 
The discourse relations between sentences are used to help order summary sentences.
In this work, we propose to incorporate explicit graph structure into the end-to-end summary decoding process. 
Graph edges are used to guide the summary generation process via a hierarchical graph attention, which is composed by a \emph{global} graph attention and a \emph{local} normalized attention.
As other components in the graph decoding layer are similar to the Transformer architecture, we focus on the extension of hierarchical graph attention. 

\paragraph{Global Graph Attention} The global graph attention is developed to capture the paragraph-level context information in the encoder part.
Different from the context attention in Transformer, we utilize the explicit graph structure to regularize the attention distributions so that graph representations of documents can be used to guide the summary generation process.

Let $y_{t}^{l-1}$ denotes the output of the $(l-1)$-th graph decoding layer for the $t$-th token in the summary. 
We assume that each token will align with several related paragraphs and one of them is at the central position.
Since the prediction of the central position depends on the corresponding query token, we apply a feed-forward network to transform $y_{t}^{l-1}$ into a positional hidden state, which is then mapped into a scalar $s_t$ by a linear projection:
\begin{equation}
\begin{aligned}
    s_t = L*sigmoid(U_{p}^{T}tanh(W_{p}y_{t}^{l-1}))
\end{aligned}
\end{equation}
\noindent where $W_{p} \in \mathbb{R}^{d*d}$ and $U_{p} \in \mathbb{R}^{d}$ denote weight matrix. 
$s_t$ indicates the central position of paragraphs that are mapped by the $t$-th summary token. 
With the central position, other paragraphs are determined by the graph structure. 
Then an attention distribution over all paragraphs under the regularization of the graph structure can be obtained:
\begin{equation}
\begin{aligned}
    \beta_{tj} =& softmax(e_{tj} -\frac{(1 - \mathbb{G}[s_t][j])^2}{2\sigma^2}) \\
\end{aligned}
\end{equation}
\noindent where $e_{tj}$ denotes the attention weight between token vector $y_{t}^{l-1}$ and paragraph vector $x_{j}$, which is computed similarly to Equation~\ref{eq3}. The global context vector can be obtained as a weighted sum of paragraph vectors: $g_t = \sum_{j=1}^{L} \beta_{tj} x_{j}$

In our decoder, graphs are also modeled as a Gaussian bias. 
Different from the encoder, a central mapping position is firstly decided and then graph relations corresponding to that position are used to regularize the attention distributions $\beta_{tj}$. 
This way, the relations in graphs are used to help align the information between source input and summary output globally, thus guiding the summary decoding process.

\paragraph{Local Normalized Attention} Then, a local normalized attention is developed to capture the token-level context information within each paragraph.
The local attention is applied 
to each paragraph independently and normalized by the global graph attention. 
This way, our model can process longer inputs effectively.

Let $\gamma_{t,ji}$ denotes the local attention distributions of the $t$-th summary token over the $i$-th token in the $j$-th input paragraph, the normalized attention is computed by:
\begin{equation}
\begin{aligned}
    \hat{\gamma}_{t,ji} = \gamma_{t,ji}\beta_{tj}
\end{aligned}
\end{equation}
\noindent and the local context vector can be computed as a weighted sum of token vectors in all paragraphs: $l_t = \sum_{j=1}^{L}\sum_{k=1}^{n} \hat{\gamma}_{t,ji}x_{ji}$

Finally, the output of the hierarchical graph attention component is computed by concatenating and linearly transforming the global and local context vector:
\begin{equation}
\begin{aligned}
    d_t = U_{d}^{T}[g_t,l_t]
\end{aligned}
\end{equation}
\noindent where $U_{d} \in \mathbb{R}^{2d*d}$ is a weight matrix. 
Through combining the local and global context, the decoder can utilize the source information more effectively.


\subsection{Combined with Pre-trained LMs}
\label{ssect:plm}
Our model can be easily combined with pre-trained LMs.
Pre-trained LMs are mostly based on sequential architectures which are more effective on short text.
For example, both BERT \citep{devlin2018bert} and RoBERTa \citep{liu2019roberta} are pre-trained with maximum 512 tokens.
\citet{liu2019text} propose to utilize BERT on single document summarization tasks.
They truncate the input documents to 512 tokens on most tasks.
However, thanks to the graph modeling, our model can process much longer inputs.
A natural idea is to combine our graph model with pretrained LMs so as to combine the advantages of them. 
Specifically, the token-level transformer encoding layer of our model can be replaced by a pre-trained LM like BERT.



In order to take full advantage of both our graph model and pre-trained LMs, the input documents are formatted in the following way:
\begin{quote}
  \small{[CLS] first paragraph [SEP] [CLS] second paragraph [SEP] \ldots [CLS] last paragraph [SEP]}
\end{quote}
\noindent Then they are encoded by a pre-trained LM, and the output vector of the ``[CLS]'' token is used as the vector of the corresponding paragraph.
Finally, all paragraph vectors are fed into our graph encoder to learn global representations.
Our graph decoder is further used to generate the summaries.

\section{Experiments}
\label{sect:exp}

\subsection{Experimental Setup}
\label{ssect:exp_set}

\paragraph{Graph Representations} We experiment with three well-established graph representations: similarity graph, topic graph and discourse graph.
The similarity graph is built based on tf-idf cosine similarities between paragraphs to capture lexical relations.
The topic graph is built based on LDA topic model \citep{blei2003latent} to capture topic relations between paragraphs.
The edge weights are cosine similarities between the topic distributions of the paragraphs.
The discourse graph is built to capture discourse relations based on discourse markers (e.g. however, moreover), co-reference and entity links as in \citet{christensen2013towards}.
Other types of graphs can also be used in our model.
In our experiments, if not explicitly stated, we use the \emph{similarity graph} by default as it has been most widely used in previous work.

\paragraph{WikiSum Dataset} We follow \citet{liu2018generating} and \citet{liu2019hierarchical} in treating the generation of lead Wikipedia sections as a MDS task.
The source documents are reference webpages of the Wikipedia article and top 10 search results returned by Google, while the summary is the Wikipedia article's first section.
As the source documents are very long and messy, they are split into multiple paragraphs by line-breaks.
Further, the paragraphs are ranked by the title and top ranked paragraphs are selected as input for MDS systems.
We directly utilize the ranking results from \citet{liu2019hierarchical} and top-40 paragraphs are used as source input.
The average length of each paragraph and the target summary are 70.1 tokens and 139.4 tokens, respectively.
For the seq2seq baselines, paragraphs are concatenated as a sequence in the ranking order, and lead tokens are used as input.
The dataset is split into 1,579,360 instances for training, 38,144 for validation and 38,205 for testing, similar to \citet{liu2019hierarchical}.
We build similarity graph representations over paragraphs on this dataset.

\paragraph{MultiNews Dataset} Proposed by \citet{fabbri2019multi}, MultiNews dataset consists of news articles and human-written summaries.
The dataset comes from a diverse set of news sources (over 1500 sites).
Different from the WikiSum dataset, MultiNews is more similar to the traditional MDS dataset such as DUC, but is much larger in scale.
As in \citet{fabbri2019multi}, the dataset is split into 44,972 instances for training, 5,622 for validation and 5,622 for testing.
The average length of source documents and output summaries are 2103.5 tokens and 263.7 tokens, respectively. 
For the seq2seq baselines, we truncate $N$ input documents to $L$ tokens by taking the first $L/N$ tokens from each source document. 
Then we concatenate the truncated source documents into a sequence by the original order.
Similarly, for our graph model, the input documents are truncated to $M$ paragraphs by taking the first $M/N$ paragraphs from each source document. 
We build all three types of graph representations on this dataset to explore the influence of graph types on MDS.

\paragraph{Training Configuration} 
We train all models with maximum likelihood estimation, and use label smoothing \citep{szegedy2016rethinking} with smoothing factor 0.1.
The optimizer is Adam \citep{kingma2014adam} with learning rate 2, $\beta_1$=0.9 and $\beta_2$=0.998.
We also apply learning rate warmup over the first 8,000 steps and decay as in \citep{vaswani2017attention}.
Gradient clipping with maximum gradient norm 2.0 is also utilized during training.
All models are trained on 4 GPUs (Tesla V100) for 500,000 steps with gradient accumulation every four steps.
We apply dropout with probability 0.1 before all linear layers in our models.
The number of hidden units in our models is set as 256, the feed-forward hidden size is 1,024, and the number of heads is 8.
The number of transformer encoding layers, graph encoding layers and graph decoding layers are set as 6, 2 and 8, respectively.
The parameter $\sigma$ is set as 2.0 after tuning on the validation dataset.
During decoding, we use beam search with beam size 5 and length penalty with factor 0.6.
Trigram blocking is used to reduce repetitions.

For the models with pretrained LMs, we apply different optimizers for the pretrained part and other parts as in \citep{liu2019text}.
Two Adam optimizers with $\beta_1$=0.9 and $\beta_2$=0.999 are used for the pretrained part and other parts, respectively.
The learning rate and warmup steps for the pretrained part are set as 0.002 and 20000, while 0.2 and 10000 for other parts.
Other model configurations are in line with the corresponding pretrained LMs. 
We choose the base version of BERT, RoBERTa and XLNet in our experiments.

\subsection{Evaluation Results}
\label{ssect:exp_res}
We evaluate our models on both the WikiSum and MultiNews datasets to validate the efficiency of them on different types of corpora. 
The summarization quality is evaluated using ROUGE $F_1$ \citep{lin2004automatic}.
We report unigram and bigram overlap (ROUGE-1 and ROUGE-2) between system summaries and gold references as a means of assessing informativeness, and the longest common subsequence (ROUGE-L\footnote{For fair comparison with previous work \citep{liu2019hierarchical, liu2018generating}, we report the summary-level ROUGE-L results on both the two datasets. The sentence-level ROUGE-L results are reported in the Appendix.}) as a means of accessing fluency.

\begin{table}[t!]
\centering
\begin{tabular}{llll}
\hline
\textbf{Model} & R-1 & R-2 & R-L \\
\hline
Lead & 38.22 & 16.85 & 26.89 \\
LexRank & 36.12 & 11.67 & 22.52 \\
\hline
FT & 40.56 & 25.35 & 34.73 \\
BERT+FT & 41.49 & 25.73 & 35.59 \\
XLNet+FT & 40.85 & 25.29 & 35.20 \\
RoBERTa+FT & 42.05 & 27.00 & 36.56 \\
T-DMCA & 40.77 & 25.60 & 34.90 \\
HT & 41.53 & 26.52 & 35.76 \\
\hline
GraphSum & 42.63 & 27.70 & 36.97\\
GraphSum+RoBERTa & \textbf{42.99} & \textbf{27.83} & \textbf{37.36} \\
\hline
\end{tabular}
\caption{\label{wikisum-exp}
Evaluation results on the WikiSum test set using ROUGE $F_1$. R-1, R-2 and R-L are abbreviations for ROUGE-1, ROUGE-2 and ROUGE-L, respectively.
}
\end{table}

\paragraph{Results on WikiSum} Table \ref{wikisum-exp} summarizes the evaluation results on the WikiSum dataset.
Several strong extractive baselines and abstractive baselines are also evaluated and compared with our models.
The first block in the table shows the results of extractive methods Lead and LexRank \citep{erkan2004lexrank}.
The second block shows the results of abstractive methods: (1) FT (Flat Transformer), a transformer-based encoder-decoder model on a flat token sequence; (2) T-DMCA, the best performing model of \citet{liu2018generating}; (3) HT (Hierarchical Transformer), a model with hierarchical transformer encoder and flat transformer decoder, proposed by \citet{liu2019hierarchical}.
We report their results following \citet{liu2019hierarchical}.
The last block shows the results of our models, which are feed with 30 paragraphs (about 2400 tokens) as input.
The results show that all abstractive models outperform the extractive ones.
Compared with FT, T-DMCA and HT, our model GraphSum achieves significant improvements on all three metrics, which demonstrates the effectiveness of our model.

\begin{table}[t!]
\centering
\begin{tabular}{llll}
\hline
\textbf{Model} & R-1 & R-2 & R-L \\
\hline
Lead & 41.24 & 12.91 & 18.84 \\
LexRank & 41.01 & 12.69 & 18.00 \\
\hline
PG-BRNN & 43.77 & 15.38 & 20.84 \\
HiMAP & 44.17 & 16.05 & 21.38 \\
FT & 44.32 & 15.11 & 20.50 \\
RoBERTa+FT & 44.26 & 16.22 & 22.37 \\
HT & 42.36 & 15.27 & 22.08 \\
\hline
GraphSum & 45.02 & 16.69 & 22.50 \\
{\small G.S.(Similarity)+RoBERTa} & 45.93 & 17.33 & 23.33 \\
{\small G.S.(Topic)+RoBERTa} & \textbf{46.07} & 17.42 & 23.21 \\
{\small G.S.(Discourse)+RoBERTa} & 45.87 & \textbf{17.56} & \textbf{23.39} \\
\hline
\end{tabular}
\caption{\label{multinews-res}
Evaluation results on the MultiNews test set. We report the \textbf{summary-level} ROUGE-L value. The results of different graph types are also compared.
}
\end{table}

Furthermore, we develop several strong baselines which combine the Flat Transformer with pre-trained LMs.
We replace the encoder of FT by the base versions of pre-trained LMs, including BERT+FT, XLNet+FT and RoBERTa+FT.
For them, the source input is truncated to 512 tokens~\footnote{Longer inputs don't achieve obvious improvements.}.
The results show that the pre-trained LMs significantly improve the summarization performance.
As RoBERTa boosts the summarization performance most significantly, we also combine it with our GraphSum model, namely GraphSum+RoBERTa~\footnote{As XLNet and BERT achieve worse results than RoBERTa, we only report the results of GraphSum+RoBERTa}.
The results show that GraphSum+RoBERTa further improves the summarization performance on all metrics, demonstrating that our graph model can be effectively combined with pre-trained LMs.
The significant improvements over RoBERTa+FT also demonstrate the effectiveness of our graph modeling even with pre-trained LMs.


\paragraph{Results on MultiNews} Table \ref{multinews-res} summarizes the evaluation results on the MultiNews dataset. 
Similarly, the first block shows two popular extractive baselines, and the second block shows several strong abstractive baselines.
We report the results of Lead, LexRank, PG-BRNN, HiMAP and FT following \citet{fabbri2019multi}.
The last block shows the results of our models.
The results show that our model GraphSum consistently outperforms all baselines, which further demonstrate the effectiveness of our model on different types of corpora.
We also compare the performance of RoBERTa+FT and GraphSum+RoBERTa, which show that our model significantly improves all metrics.

The above evaluation results on both WikiSum and MultiNews dataset both validate the effectiveness of our model.
The proposed method to modeling graph in end-to-end neural model greatly improves the performance of MDS.

\subsection{Model Analysis}
\label{ssect:exp_any}
We further analyze the effects of graph types and input length on our model, and validate the effectiveness of different components of our model by ablation studies.

\begin{table}
\centering
\begin{tabular}{lllll}
\hline
\textbf{Len} & \textbf{Model} & R-1 & R-2 & R-L \\
\hline
\multirow{3}{*}{500} 
& HT & 41.08 & 25.83 & 35.25 \\
& GraphSum & \textbf{41.55} & \textbf{26.24} & \textbf{35.59} \\
& $\nabla$ & +0.47 & +0.41 & +0.34 \\
\hline
\multirow{3}{*}{800} 
& HT & 41.41 & 26.46 & 35.79 \\
& GraphSum & \textbf{41.70} & \textbf{26.87} & \textbf{36.10} \\
& $\nabla$ & +0.29 & +0.41 & +0.31 \\
\hline
\multirow{3}{*}{1600} 
& HT & 41.53 & 26.52 & 35.76 \\
& GraphSum & \textbf{42.48} & \textbf{27.52} & \textbf{36.66} \\
& $\nabla$ & +0.95 & +1.00 & +0.90 \\
\hline
\multirow{3}{*}{2400} 
& HT & 41.68 & 26.53 & 35.73 \\
& GraphSum & \textbf{42.63} & \textbf{27.70} & \textbf{36.97} \\
& $\nabla$ & +0.95 & +1.17 & +1.24 \\
\hline
\multirow{3}{*}{3000} 
& HT & 41.71 & 26.58 & 35.81 \\
& GraphSum & \textbf{42.36} & \textbf{27.47} & \textbf{36.65} \\
& $\nabla$ & +0.65 & +0.89 & +0.84 \\
\hline
\end{tabular}
\caption{\label{len-exp}
Comparison of different input length on the WikiSum test set using ROUGE $F_1$. 
$\nabla$ indicates the improvements of GraphSum over HT.
}
\end{table}


\paragraph{Effects of Graph Types} To study the effects of graph types, the results of GraphSum+RoBERTa with similarity graph, topic graph and discourse graph are compared on the MultiNews test set.
The last block in Table \ref{multinews-res} summarizes the comparison results, which show that the topic graph achieves better performance than similarity graph on ROUGE-1 and ROUGE-2, and the discourse graph achieves the best performance on ROUGE-2 and ROUGE-L.
The results demonstrate that graphs with richer relations are more helpful to MDS.

\paragraph{Effects of Input Length} Different lengths of input may affect the summarization performance seriously for Seq2Seq models, so most of them restrict the length of input and only feed the model with hundreds of lead tokens.
As stated by \citet{liu2019hierarchical}, the FT model achieves the best performance when the input length is set to 800 tokens, while longer input hurts performance.
To explore the effectiveness of our GraphSum model on different length of input, we compare it with HT on 500, 800, 1600, 2400 and 3000 tokens of input respectively.
Table \ref{len-exp} summarizes the comparison results, which show that our model outperforms HT on all length of input.
More importantly, the advantages of our model on all three metrics tend to become larger as the input becomes longer.
The results demonstrate that modeling graph in the end-to-end model enables our model process much longer inputs with better performance.

\begin{table}
\centering
\begin{tabular}{llll}
\hline
\textbf{Model} & Rouge-1 & Rouge-2 & Rouge-L \\
\hline
GraphSum & \textbf{42.63} & \textbf{27.70} & \textbf{36.97}\\
w/o graph dec & 42.06 & 27.13 & 36.33 \\
w/o graph enc & 40.61 & 25.90 & 35.26 \\
\hline
\end{tabular}
\caption{\label{albation-exp}
Ablation study on the WikiSum test set.
}
\end{table}

\paragraph{Ablation Study} Table \ref{albation-exp} summarizes the results of ablation studies aiming to validate the effectiveness of individual components.
Our experiments confirmed that incorporating well-known graphs into the encoding process by our graph encoder (see w/o graph enc) and utilizing graphs to guide the summary decoding process by our graph decoder (w/o graph dec) are both beneficial for MDS.

\subsection{Human Evaluation}
\label{ssect:exp_hum}

In addition to the automatic evaluation, we also access system performance by human evaluation.
We randomly select 50 test instances from the WikiSum test set and 50 from the MultiNews test set, and invite 3 annotators to access the outputs of different models independently.
Annotators access the overall quality of summaries by ranking them taking into account the following criteria: (1) \emph{Informativeness}: does the summary convey important facts of the input? (2) \emph{Fluency}: is the summary fluent and grammatical? (3) \emph{Succinctness}: does the summary avoid repeating information?
Annotators are asked to ranking all systems from 1(best) to 5 (worst). 
Ranking could be the same for different systems if they have similar quality.
For example, the ranking of five systems could be 1, 2, 2, 4, 5 or 1, 2, 3, 3, 3.
All systems get score 2, 1, 0, -1, -2 for ranking 1, 2, 3, 4, 5 respectively.
The rating of each system is computed by averaging the scores on all test instances. 

Table \ref{human-exp} summarizes the comparison results of five systems. 
Both the percentage of ranking results and overall ratings are reported.
The results demonstrate that GraphSum and GraphSum+RoBERTa are able to generate higher quality summaries than other models.
Specifically, the summaries generated by GraphSum and GraphSum+RoBERTa usually contains more salient information, and are more fluent and concise than other models.
The human evaluation results further validates the effectiveness of our proposed models.

\begin{table}[!t]
\centering
\small
\begin{tabular}{lllllll}
\hline
\textbf{Model} & 1 & 2 & 3 & 4 & 5 & Rating \\
\hline
FT & 0.18 & 0.21 & 0.23 & 0.16 & 0.22 & -0.03$^*$ \\
R.B.+FT & 0.32 & 0.22 & 0.17 & 0.19 & 0.10 & 0.49$^*$ \\
HT & 0.21 & 0.32 & 0.12 & 0.15 & 0.20 & 0.19$^*$ \\
GraphSum & 0.42 & 0.30 & 0.17 & 0.10 & 0.01 & 1.02 \\
G.S.+R.B. & 0.54 & 0.24 & 0.10 & 0.08 & 0.04 & \textbf{1.16} \\
\hline
\end{tabular}
\caption{
\label{human-exp}
Ranking results of system summaries by human evaluation.
1 is the best and 5 is the worst. 
The larger rating denotes better summary quality.
R.B. and G.S. are the abbreviations of RoBERTa and GraphSum, respectively. 
$^*$ indicates the overall ratings of the corresponding model are significantly (by Welch’s t-test with $p<0.01$) outperformed by our models GraphSum and GraphSum+RoBERTa.}
\end{table}

\section{Conclusion}
\label{sect:con}
In this paper we explore the importance of graph representations in MDS and propose to leverage graphs to improve the performance of neural abstractive MDS.
Our proposed model is able to incorporate explicit graph representations into the document encoding process to capture richer relations within long inputs, and utilize explicit graph structure to guide the summary decoding process to generate more informative, fluent and concise summaries.
We also propose an effective method to combine our model with pre-trained LMs, which further improves the performance of MDS significantly.
Experimental results show that our model outperforms several strong baselines by a wide margin. 
In the future we would like to explore other more informative graph representations such as knowledge graphs, and apply them to further improve the summary quality.

\section*{Acknowledgments}
This work was supported by the National Key Research and Development Project of China (No. 2018AAA0101900).

\bibliography{acl2020-cameraready-0502}
\bibliographystyle{acl2020-cameraready-0502}

\appendix

\section{Appendix}
We report the sentence-level ROUGE-L evaluation results of our models on both the two datasets, so that future work can compare with them conveniently.

\begin{table}[h]
\centering
\begin{tabular}{llll}
\hline
\textbf{Model} & R-1 & R-2 & R-L \\
RoBERTa+FT & 42.05 & 27.00 & 40.05 \\
GraphSum & 42.63 & 27.70 & 40.13\\
GraphSum+RoBERTa & \textbf{42.99} & \textbf{27.83} & \textbf{40.97} \\
\hline
\end{tabular}
\caption{\label{wikisum-exp}
Evaluation results on the WikiSum test set with \textbf{sentence-level} ROUGE-L value.
}
\end{table}

\begin{table}[!t]
\centering
\begin{tabular}{llll}
\hline
\textbf{Model} & R-1 & R-2 & R-L \\
\hline
RoBERTa+FT & 44.26 & 16.22 & 40.64 \\
GraphSum & 45.02 & 16.69 & 41.11 \\
{\small G.S.(Similarity)+RoBERTa} & 45.93 & 17.33 & 42.02 \\
{\small G.S.(Topic)+RoBERTa} & \textbf{46.07} & 17.42 & \textbf{42.22} \\
{\small G.S.(Discourse)+RoBERTa} & 45.87 & \textbf{17.56} & 42.00 \\
\hline
\end{tabular}
\caption{\label{multinews-res}
Evaluation results on the MultiNews test set with \textbf{sentence-level} ROUGE-L value.
}
\end{table}

\end{document}